\documentclass[12pt]{article}

\usepackage{sbc-template}
\usepackage{graphicx,url}
\usepackage[utf8]{inputenc}
\usepackage[brazil]{babel}
\usepackage{multirow}
\usepackage{adjustbox}

\usepackage{todonotes}
\usepackage{nicefrac}

\title{PROPOE 2: Avanços na Síntese Computacional de Poemas Baseados em Prosa Literária Brasileira}

\author{Felipe José D. Sousa \inst{1}, Sarah P. Cerqueira\inst{1}, João Queiroz\inst{2}, Angelo Loula \inst{1}  }

\address{Universidade Estadual de Feira de Santana (UEFS)\\ 
    Feira de Santana -- BA -- Brasil
\nextinstitute
Universidade Federal de Juiz de Fora (UFJF)\\ 
    Juiz de Fora -- MG -- Brasil
    \email{\{felipedamasceno97,cerqueira.sarahp,queirozj\}@gmail.com, angelocl@uefs.br}
}

\begin{document} 

\maketitle

\begin{abstract}
The computational generation of poems is a complex task, which involves several sound, prosodic and rhythmic resources. In this work we present PROPOE 2, with the extension of structural and rhythmic possibilities compared to the original system, generating poems from metered sentences extracted from the prose of Brazilian literature, with multiple rhythmic assembly criteria. These advances allow for a more coherent exploration of rhythms and sound effects for the poem. Results of poems generated by the system are demonstrated, with variations in parameters to exemplify generation and evaluation using various criteria.
\end{abstract}
     
\begin{resumo} 

A geração computacional de poemas é uma tarefa complexa, que envolve diversos recursos sonoros, prosódicos e rítmicos.  Neste trabalho apresentamos PROPOE 2, com a ampliação de possibilidades estruturais e rítmicas em relação ao sistema original, gerando poemas a partir de sentenças metrificadas extraídas da prosa da literatura brasileira, com múltiplos critérios rítmicos de montagem. Esses avanços permitem uma exploração mais coerente de ritmos e efeitos sonoros para o poema. Resultados de poemas gerados pelo sistema são demonstrados, com variações de parâmetros para exemplificar a geração e a avaliação pelos variados critérios.
\end{resumo}

\section{Introdução}

A geração computacional de poemas é uma tarefa complexa, e depende da modelagem de diferentes níveis de organização da linguagem -- som, ritmo, sintaxe, morfologia, semântica \cite{gonccalo2024automatic}. Há projetos de desenvolvimento de métodos computacionais para geração de poemas desde, ao menos, a década de 1960. Mas foi somente na virada do milênio que essa tarefa passou a atrair a atenção da comunidade científica de Ciência da Computação \cite{oliveira2017survey}.

Usualmente, considera-se um poema uma classe ou categoria de texto verbal organizado em linhas (versos), e conjunto de linhas (estrofes de versos), com notável presença de dispositivos e recursos visuais, sonoros, prosódicos, rítmicos \cite{goldstein1985versos}. A composição do poema se baseia em propriedades estruturais que podem incluir o metro, a rima, e a distribuição temporal de pulsos e intervalos (ritmo). O metro é a medida correspondente à quantidade de sílabas poéticas dos versos \cite{estudoanalitico}. Em português, a contagem do metro termina na última sílaba poética tônica, que nem sempre correspondem às sílabas gramaticais  \cite{ goldstein1985versos}. Os elementos sonoros do poema estão intimamente ligados ao fenômeno dominante do ritmo \cite{estudoanalitico}, formado pela alternância entre sílabas acentuadas e não-acentuadas, ou entre sílabas constituídas de vogais longas e breves  \cite{ goldstein1985versos}. A rima é outro importante dispositivo para obtenção de efeitos de sonoridade no verso \cite{estudoanalitico}, e pode ser definida como a similitude de sons na terminação das palavras \cite{bandeira1997versificaccao}. Ela pode ocorrer externamente, quando a repetição de sons similares acontece no final dos versos  \cite{ goldstein1985versos}, e internamente, quando a rima ocorre dentro de um verso, ou quando se distribui entre o final de um verso e o meio do verso seguinte [Bandeira 1997].


Considerando estas propriedades (rítmicas e fonológicas), o sistema PROPOE 
PROPOE (\textit{Prose to Poem}) \cite{azevedo2018sintese} 
gera poemas com diversos tipos de rima, e ritmos, a partir de um conjunto de sentenças metrificadas extraídas da prosa literária. A mineração de sentenças metrificas é realizada pela ferramenta computacional  
MIVES (\textit{Mining Verse Structure}) \cite{carvalho28identificaccao}
. Ela realiza a escansão de segmentos frásicos de textos literários e classifica as estruturas encontradas conforme padrões métricos e rítmicos normatizados. O PROPOE recebe como entrada um conjunto de sentenças metrificadas e a partir de parâmetros relacionados à estrutura do poema e aos critérios rítmicos, que guiam um ``\textit{algoritmo} guloso" na seleção das sentenças candidatas e na estratégia de montagem.

Em sua primeira versão, o PROPOE apresenta um conjunto reduzido de variações da estrutura do poema, do metro dos versos e de critérios rítmicos para seleção individual de sentenças. Neste trabalho, descrevemos o PROPOE 2, que inclui a ampliação de possibilidades estruturais e rítmicas na geração de poemas. O sistema prevê maior flexibilidade na variação dos parâmetros estruturais do poema, ajuste na estratégia de filtragem de sentenças candidatas, novos critérios de seleção de sentenças, com maior diversidade de ritmos e efeitos sonoros, e com possibilidade de ponderação entre múltiplos critérios na montagem do poema.

\section{Trabalhos relacionados}


Para lidar com a complexidade de gerar poemas automaticamente, muitos estudos restringem tantos os aspectos poéticos abordados quanto as características linguísticas do texto. Na  \cite{gervas2001expert}, os poemas em espanhol são gerados utilizando uma técnica de raciocínio baseado em casos (\textit{Case-Based Reasoning} - CBR). É possível criar poemas nos formatos de romance (uma estrofe em que todas as linhas rimam), quarteto (uma estrofe com quatro versos) ou terceto encadenado (todas as estrofes com três versos). O formato é selecionado internamente pela ferramenta para se adequar aos parâmetros de número de versos e grau de formalidade, que pode ser informal (oito sílabas) ou formal (onze sílabas). Em \cite{manurung2012}, foi proposta a aplicação McGONAGALL, que gera poemas em inglês mediante um modelo de busca estocástica, utilizando um algoritmo evolutivo. Nessa aplicação, os autores concentraram-se mais nos aspectos linguísticos de sintaxe e semântica, sendo o metro a única característica poética possível de ser gerenciada.

No  PoeTryMe \cite{oliveira2012poetryme}, a construção de poemas é realizada via versos gerados por meio de uma rede semântica e outra gramatical, que interagem através de uma estratégia de geração, a qual pode selecionar versos de forma aleatória, utilizando o método \textit{generate-and-test} ou um algoritmo evolutivo. A ferramenta compõe poemas no idioma português com estruturas variadas, que pode ser escolhida pelo usuário ao selecionar um determinado template. Um template define o número de estrofes, versos por estrofe e as configurações de metro.

Entre as pesquisas que geram poemas clássicos chineses, o formato mais comum é a quadra (quatro linhas) com cinco ou sete caracteres. Em \cite{yan2013poet}, a criação de quadras é tratada como um problema de otimização, baseado em uma estrutura de sumarização generativa. O modelo PPG (\textit{Planning-based Poetry Generation}) \cite{wang2016} compõe quadras utilizando um framework de atenção baseado em uma RNN (\textit{Recurrent Neural Network}) \textit{encoder-decoder} modificada. Já no KPG (\textit{Keyword Team Poetry Generation}) \cite{gao2020} que também gera quadras chinesas, o modelo é formado por redes do tipo LSTM (\textit{Long Short-Term Memory}).

A aplicação PoeLM \cite{ormazabal2022poelm} cria diferentes tipos de poemas em espanhol e basco, com base em textos não poéticos, dado um esquema de metro e rima. O PoeLM gera poemas candidatos através de uma \textit{transformer} LM (\textit{Language Model}), que posteriormente são filtrados por nem todos seguirem as restrições passadas.   De maneira semelhante, em \cite{van2020automatic}, os poemas também são elaborados a partir de textos não poéticos, para os idiomas inglês e francês. As sentenças candidatas a compor o poema são geradas por um modelo RNN \textit{encoder-decoder}, sendo incorporadas ao poema por meio de um framework de otimização geral. No artigo, os formatos de poema que a aplicação pode gerar não são explicitados, mas no código disponibilizado no GitHub estão especificadas estruturas como soneto, quadra, pantum formado por quatro quadras e outra composta por duas quadras.
 

No PROPOE, a criação de poemas é abordada como um problema de otimização, considerando múltiplos critérios rítmicos \cite{azevedo2018sintese}. Utilizando um ``algoritmo guloso", sentenças metrificadas extraídas de obras literárias são filtradas e combinadas para maximizar critérios rítmicos. Diferente da maioria das aplicações de geração automática de poemas, o PROPOE, em sua primeira versão, reutiliza senteças da prosa literária, com seus potenciais paralelismos estruturais, e permite a aplicação de variados critérios rítmicos e fonológicos na otimização para a montagem de um poema. Além disso, na literatura, não foram encontradas outras ferramentas capazes de gerar poemas em português, além do PROPOE (Brasil) e do PoeTryMe (Portugal).

\section{Metodologia}

\begin{figure}[ht]
\centering
\includegraphics[scale=0.45]{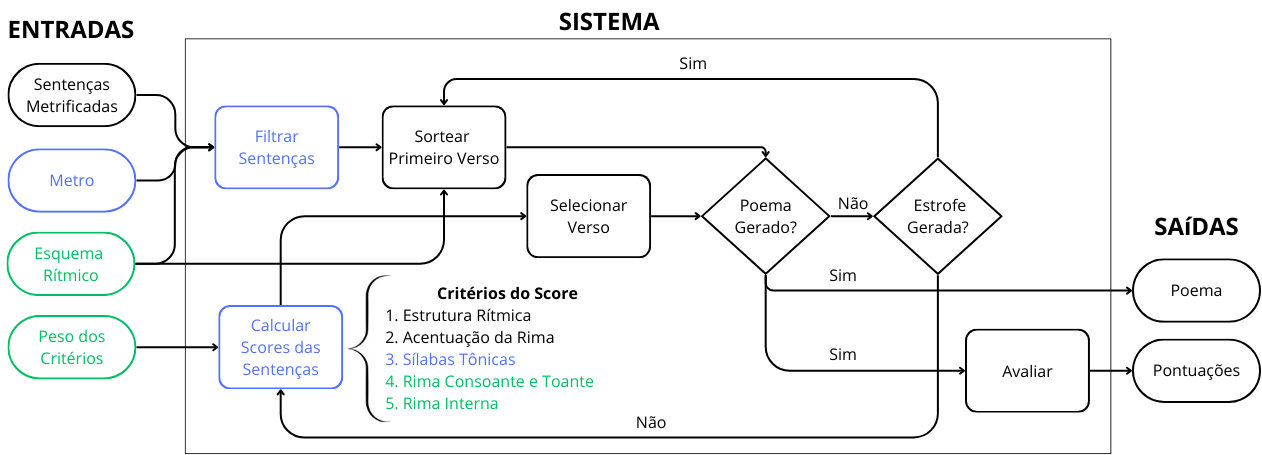}
\caption{Sistema PROPOE 2. Os blocos com letra preta representam o que foi mantido da versão original, em azul o que foi alterado e em verde o que foi adicionado.}
\label{figura:sistema}

\end{figure}

Neste trabalho, é proposto o PROPOE 2, que amplia a integração de aspectos rítmicos na seleção dos versos do poema e oferece maior variabilidade na escolha estrutural e rítmica para o poema. Partindo da primeira versão do PROPOE, vários aspectos foram ajustados e ampliados: definição flexível da estrutura de estrofes, versos e rimas,  especificação individual do metro de cada verso, ajuste na filtragem de segmentos, ajuste na aplicação de critérios para seleção de versos, e ajuste na avaliação de sílabas tônicas. Ademais, foram adicionados novos critérios de avaliação de versos candidatos, incluindo o esquema rítmico e os pesos para ponderar múltiplos  critérios de seleção dos versos. Também foram inseridos os critérios de rima interna, consoante e toante. A Figura \ref{figura:sistema} apresenta o diagrama do PROPOE 2, evidenciando suas mudanças em relação à primeira versão.

\subsection{Entradas}

A principal entrada do PROPOE são um conjunto de sentenças metrificadas obtidas pela ferramenta MIVES \cite{carvalho28identificaccao}, que realiza a escansão de sentenças completas e parciais de textos de obras da literatura brasileira. Para cada sentença estão disponíveis a versão original, suas versões escandidas e os metros associados, uma vez que o processo de escansão, considerando fenômenos fonológicos, pode ter múltiplas variações.

O parâmetro esquema rítmico é uma representação da estrutura de estrofes do poema e de rimas dos versos. Por exemplo, o esquema rítmico ``AAB BBA" representa um poema com duas estrofes, em que cada estrofe tem três versos. Os versos um, dois e seis devem ter a mesma rima, pois são representados pela mesma letra ``A". Da mesma forma, os versos três, quatro e cinco, representados pela letra ``B", também devem rimar entre si. A rima considerada é a rima externa que corresponde a sílaba final do verso, então dois versos com a mesma letra são versos com a mesma sílaba final.

O metro é uma entrada opcional, assim como na versão original do sistema. No entanto, foi alterada para permitir a escolha de um metro específico para cada verso, em vez de um único metro para todos os versos, aumentando assim a flexibilidade dos poemas que podem ser gerados. Caso o metro de um verso não seja escolhido, o metro é definido de forma aleatória.

Os pesos dos critérios permitem determinar a importância de cada critério rítmico utilizado na seleção dos versos que irão compor o poema. Os pesos escolhidos são multiplicados pela pontuação dos critérios durante a montagem do poema. Os critérios possuem pontuação com valores entre 0 e 1, e o peso é aplicado para cada pontuação para compor a pontuação final por média ponderada.

\subsection{Filtragem das Sentenças Métricas}

A geração dos poemas inicia com a importação das sentenças métricas, que são, em seguida, agrupados com base em sua última sílaba poética, resultando em grupos de sentenças que rimam entre si. Por sua vez, cada grupo é organizado em subgrupos conforme o metro das sentenças.

A filtragem das sentenças visa manter somente os grupos de sentenças que atendem aos requisitos de metro e esquema rítmico requeridos. Então, é verificado se os grupos têm quantidade suficiente de versos para atender aos requisitos do poema. Considerando um exemplo onde o esquema rítmico é ``AA BBA" e o metro dos primeiros dois versos é 9 e o dos últimos três é 10, um grupo só poderia ser utilizado para compor os versos ``A" se houvesse pelo menos dois versos com metro 9 e um com metro 10. Já para os versos ``B" seria necessário um grupo que pelo menos tivesse dois versos com metro 10. Respeitando essa restrição, são selecionados os possíveis grupos de sentença para cada letra presente no esquema rítmico. 

No PROPOE 2, os versos que compartilham a mesma palavra final dentro dos subgrupos relacionados ao metro são considerados duplicados. Por isso, ao verificar se um grupo está habilitado para compor o poema, esses versos são contabilizados como um só. Essa restrição também é aplicada durante a composição do poema: uma vez que um verso é escolhido, todos os outros com a mesma palavra final são descartados como candidatos para os demais versos. O propósito dessa condição é mitigar a tendência de selecionar versos com palavras repetidas, especialmente na última posição, um problema observado na primeira versão do PROPOE.

A etapa de filtragem finaliza atribuindo aleatoriamente uma sílaba final para cada letra presente no esquema rítmico. Este sorteio ocorre apenas entre os grupos habilitados para aquela letra, e um grupo não pode ser sorteado duas vezes para letras diferentes.

\subsection{Montagem do Poema}

A montagem do poema consiste na iteração pelas letras do esquema rítmico do primeiro ao último. Para cada letra é selecionado um verso, dentre aqueles pertencentes ao grupo de sentenças definido na etapa anterior. Caso seja o primeiro da estrofe, este é sorteado de forma aleatória e define uma referência rítmica para os próximos versos. Os demais versos são selecionados com base em similaridade rítmica com os versos anteriores, buscando aquele de maior escore, a pontuação total. Este procedimento segue verso a verso até o poema ser gerado conforme especificação.

\subsection{Escore das Sentenças Métricas}

O cálculo do escore serve como indicador de similaridade rítmica ao comparar as sentenças candidatas disponíveis do grupo com os versos do poema. Na primeira versão do PROPOE, o usuário podia somente escolher se verso de referência seria o primeiro da estrofe ou o imediatamente anterior, e se o cálculo do escore da sentença seria baseada na posição das sílabas tônicas (padrão rítmico) ou na acentuação (oxítona, paroxítona ou proparoxítona) das palavras que rimam.  Caso o valor escore empatasse entre duas ou mais sentenças, o outro critério era adicionado como desempate. 

No PROPOE 2, o escore das sentenças considera, além dos critérios da versão original, três novos critérios: igualdade de sílabas tônicas, rima interna e  rima consoante e toante. Cada um dos cinco critérios possui pontuação entre 0 e 1, e o escore de cada sentença é dado pela média ponderada da pontuação dos critérios em relação aos pesos fornecidos como parâmetro de entrada.

Na abordagem atual da versão dois, tanto o primeiro verso de uma determinada rima é tomado em consideração para calcular os critérios de acentuação da rima e rima toante e consoante, como o verso imediatamente anterior para calcular os outros critérios.  Quando o verso é o primeiro representante de uma rima, então os critérios da acentuação e rima toante e consoante não podem ser apurados, logo são desconsiderados no cômputo do escore. Essa abordagem permite que os versos tenham maior similaridade rítmica entre si, tornando o poema final mais coerente no ritmo poético.

\subsubsection{Padrão Rítmico}

O som tônico repetido na mesma posição em cada verso gera ritmo ao poema. O critério de padrão rítmico ou similaridade rítmica computa justamente a similaridade entre as posições das sílabas tônicas nos versos através do coeficiente de Jaccard (equação \ref{jaccard}).

 Para esta avaliação, as posições das sílabas tônicas em cada verso formam um conjunto composto pelos números destas posições. Considerando dois conjuntos finitos de padrões rítmos de dois versos, o coeficiente de Jaccard na equação \ref{jaccard} mede a similaridade através da razão entre a cardinalidade da interseção e união dos conjuntos.
 
\vspace{-10pt}
\begin{equation}
    Jaccard(U,V) = \frac{|U \cap V|}{|U \cup V|}
    \label{jaccard}
\end{equation}
\vspace{-15pt}

\subsubsection{Sílabas Tônicas}

Originalmente, o critério de sílabas tônicas pontuava apenas se o verso de referência e o verso candidato tivessem as sílabas tônicas nas mesmas posições, uma vez que o efeito sonoro ganha mais evidência nessa condição. Entretanto, mesmo quando as sílabas tônicas estão em posições distintas, ainda há um efeito sonoro pela repetição fonética enfatizada. Por isso, esse critério foi alterado para considerar também as sílabas tônicas em posições diferentes.

O cálculo desse critério rítmico acontece considerando ambas possibilidades. Na primeira parte, conta-se a quantidade de sílabas tônicas iguais (independente de posição) entre o verso de referência e a sentença candidata. Esse valor é dividido pela menor quantidade de sílabas tônicas entre os dois. Por exemplo, se o verso de referência tem 4 sílabas tônicas e a sentença metrificada possui 3 sílabas tônicas, sendo apenas uma sílaba igual, então o resultado é a divisão de 1 por 3, pois temos apenas uma sílaba igual e a menor quantidade de tônicas é 3. A segunda parte é semelhante à primeira, mas no numerador da divisão considera-se a quantidade de sílabas tônicas iguais que estão na mesma posição. Por fim, o resultado do critério é a média do resultado das duas partes.

\vspace{-15pt}
\subsubsection{Rima Consoante e Toante}

A correspondência sonora na rima é definida pelo critério de rima consoante e toante. A rima consoante é calculada verificando se as letras da última palavra da sentença, a partir da vogal tônica, são iguais às do verso de referência. Se forem iguais, há consoância então o critério pontua o verso com 1.0. Caso contrário, analisa-se se a rima é toante, se os versos possuem a mesma vogal tônica na última palavra (assoância), e neste o critério retorna 0,5. Se não for consoante nem toantes, critério tem pontuação zero.

\vspace{-15pt}
\subsubsection{Acentuação da Rima}
\vspace{-5pt}

As rimas podem ser classificadas quanto à acentuação. Se as palavras que rimam forem oxítonas, a rima é aguda, se forem paroxítonas, a rima é grave, se forem proparoxítonas, a rima é esdrúxula. \cite{bandeira1997versificaccao, goldstein1985versos}. Nesse critério, se a acentuação da rima for igual entre o verso de referência e a sentença candidata, a pontuação é 1.0; caso contrário, é zero.

\vspace{-15pt}
\subsubsection{Rima interna}
\vspace{-5pt}

Diferente das outras rimas, a rima interna (RI) não requer um verso de referência. Esse critério busca selecionar sentenças que possuem efeitos sonoros nas sílabas poéticas internas da sentença, por meio da repetição de sons.  A rima interna é estimada pela equação $RI = 1 - \nicefrac{SU}{ST}$, onde SU presenta a quantidade de sílabas poéticas únicas na sentença e ST é a quantidade total de sílabas poéticas. Assim, se uma sílaba se repete várias vezes na sentença, a quantidade de sílabas únicas diminui, resultando em uma pontuação maior para essa sentença neste critério.

\subsection{Avaliação final}

Após a geração do poema, o PROPOE realiza uma avaliação do poema obtido produzindo um escore geral, com objetivo de resumir as informações de todo o poema e servir de métrica para comparar com outros poemas.

Na versão inicial, a avaliação global do poema se fundamentava em quatro critérios: similaridade rítmica, sílabas tônicas, acentuação da rima e similaridade entre as palavras que rimam. Cada critério era normalizado em uma escala de [0 a 1], resultando em uma avaliação final que era a soma desses critérios. Portanto, a avaliação final de um poema podia variar entre 0 e 4.0.

Atualmente, na análise geral do poema, é gerada uma pontuação global de todo poema para cada um dos critérios calculados para as sentenças candidatas. A pontuação de cada critério do poema é formada pela média da pontuação do critério para os versos. De maneira similar, o escore global do poema é a média dos escore dos versos.

\section{Resultados}

Para demonstrar os resultados obtidos pelo PROPOE 2, foram gerados cinco poemas com variações dos parâmetros, a partir de sentenças metrificadas extraídas da obra literária brasileira \textit{Os Sertões}, de Euclides da Cunha. A Tabela \ref{parametros} apresenta as configurações utilizadas para cada um dos poemas gerados. Na coluna ``Peso dos Critérios",  são indicados os pesos atribuídos aos critérios de esquema rítmico (ER), de sílaba tônica (ST), de acentuação da rima (AC), de rima interna (RI) e de rima toante e consoante (RTC). Na coluna ``Metro" um único valor indica que todos os versos tem a mesma metrificação, caso contrário é discriminado o metro para cada verso. 

A semente aleatória fixa, na  Tabela \ref{parametros}, indica que o PROPOE irá gerar o mesmo poema desde que se mantenha os parâmetros de entrada iguais. Essa estratégia foi tomada na geração dos poemas 1, 2 e 3 para que fosse possível verificar o impacto da mudança dos parâmetros de metro e pesos dos critérios, a partir dos mesmos grupos de sentenças candidatas. 

\begin{table*}[ht]
\caption{Configuração dos poemas gerados através PROPOE 2.}
\label{parametros}
\begin{center}
\resizebox{\columnwidth}{!}{%
\begin{tabular}{c|c|c|c|c|c|c|c|c}
\hline
\multirow{2}{*}{\textbf{Poema}} & \multirow{2}{*}{\textbf{Esquema Rítmico}}  & \multirow{2}{*}{\textbf{Metro}}  &
 \multicolumn{5}{c|}{\textbf{Peso dos Critérios}} & \multirow{2}{*}{\textbf{Semente Aleatória}} \\ 

\cline{4-8}
 & & & ER & ST & AC & RI & RTC \\

\hline
 1 & AABB & 10 & 1 & 1 & 1 & 1 & 1 & Fixa \\
 \hline
  2 & AABB & 10 & 0 & 1 & 1 & 2 & 0 & Fixa \\
 \hline
  3 & AABB & 10 9 9 10 & 1 & 1 & 1 & 1 & 1 & Fixa \\
 \hline
  4 & AABB & 10 & 1 & 1 & 1 & 1 & 1 & Livre \\
 \hline
  5 & ABAB ABAB CDC CDC & 10 & 1 & 1 & 1 & 1 & 1 & Livre \\
 \hline
\end{tabular}
}

\end{center}
\end{table*}

Na Tabela \ref{poema1} é exibido o  poema 1, junto com sua versão escandida, o metro de cada verso, a posição das sílabas tônicas, bem como os valores dos critérios considerados para compor o escore e o próprio escore.  Na coluna da tabela que apresenta a escansão, o símbolo ``\//" indica a separação das sílabas poéticas e ``\#" indica a posição da sílaba tônica.

O primeiro poema gerado (Tabela \ref{poema1}) possuí uma estrofe e quatro versos com metrificação 10, conforme o esquema rítmico solicitado (Tabela \ref{parametros}). O primeiro verso rima com o segundo através da última sílaba ``te" e o terceiro rima com o quarto através da sílaba ``de".

O símbolo ``-" na Tabela \ref{poema1} representa que o critério não foi calculado para aquele verso. No primeiro verso isso acontece por ele ter sido escolhido de forma aleatória, e no terceiro verso, porque ele é o primeiro verso com rima ``B", logo, os critérios de acentuação e rima toante e consoante não possuem um verso de referência para usar no cálculo.

\begin{table*}[ht]
\caption{Primeiro poema gerado pelo PROPOE 2. }
\centering
\resizebox{\columnwidth}{!}{%

\begin{tabular}{lllcccccc}
\hline
\multicolumn{9}{l}{\multirow{4}{*}{\begin{tabular}[c]{@{}l@{}}Renovou-se a investida febrilmente\\ Da Favela batiam-nos de frente\\ Quase tudo está seco de sede\\ A tropa combalida abalou à tarde\end{tabular}}}                                                                                                    \\
\multicolumn{9}{l}{}                                                                                                                                                                                                                                                                                                  \\
\multicolumn{9}{l}{}                                                                                                                                                                                                                                                                                                  \\
\multicolumn{9}{l}{}                                                                                                                                                                                                                                                                                                  \\ \hline
\multicolumn{1}{|l|}{Escansão}                                            & \multicolumn{1}{l|}{Metro} & \multicolumn{1}{l|}{Tônicas}        & \multicolumn{1}{c|}{ER}    & \multicolumn{1}{c|}{ST}    & \multicolumn{1}{c|}{AC} & \multicolumn{1}{c|}{RI}    & \multicolumn{1}{c|}{RTC} & \multicolumn{1}{c|}{Escore} \\ \hline
\multicolumn{1}{|l|}{Re/no/v\#ou/se a in/ves/t\#i/da/ fe/bril/m\#en/te.}  & \multicolumn{1}{l|}{10}    & \multicolumn{1}{l|}{3, 6, 10}       & \multicolumn{1}{c|}{-}     & \multicolumn{1}{c|}{-}     & \multicolumn{1}{c|}{-}  & \multicolumn{1}{c|}{-}     & \multicolumn{1}{c|}{-}   & \multicolumn{1}{c|}{-}     \\ \hline
\multicolumn{1}{|l|}{da/ Fa/v\#e/la/ ba/t\#i/am/nos/ de/ fr\#en/te.}      & \multicolumn{1}{l|}{10}    & \multicolumn{1}{l|}{3, 6, 10}       & \multicolumn{1}{c|}{1}     & \multicolumn{1}{c|}{0.333} & \multicolumn{1}{c|}{1}  & \multicolumn{1}{c|}{0}     & \multicolumn{1}{c|}{1}   & \multicolumn{1}{c|}{0.667} \\ \hline
\multicolumn{1}{|l|}{Qu\#a/se/ t\#u/do/ es/t\#á/ s\#e/co/ de/ s\#e/de...} & \multicolumn{1}{l|}{10}    & \multicolumn{1}{l|}{1, 3, 6, 7, 10} & \multicolumn{1}{c|}{0.6}   & \multicolumn{1}{c|}{0}     & \multicolumn{1}{c|}{-}  & \multicolumn{1}{c|}{0.273} & \multicolumn{1}{c|}{-}   & \multicolumn{1}{c|}{0.291} \\ \hline
\multicolumn{1}{|l|}{A/ tr\#o/pa/ com/ba/l\#i/da a/ba/l\#ou à/ t\#ar/de.} & \multicolumn{1}{l|}{10}    & \multicolumn{1}{l|}{2, 6, 9, 10}    & \multicolumn{1}{c|}{0.343} & \multicolumn{1}{c|}{0}     & \multicolumn{1}{c|}{1}  & \multicolumn{1}{c|}{0.091} & \multicolumn{1}{c|}{0}   & \multicolumn{1}{c|}{0.287} \\ \hline
\multicolumn{3}{|c|}{Avaliação}                                                                                                              & \multicolumn{1}{c|}{0.648}  & \multicolumn{1}{c|}{0.111}  & \multicolumn{1}{c|}{1}  & \multicolumn{1}{c|}{0.121}  & \multicolumn{1}{c|}{0.5} & \multicolumn{1}{c|}{0.415}  \\ \hline

\multicolumn{9}{l}{
\footnotesize{ (Tônicas informa a posição das sílabas tônicas, ER representa o resultado do critério do esquema rítmico, ST é da sílaba tônica, AC é}
}\\
\multicolumn{9}{l}{
\footnotesize{ acentuação, RI é rima interna e RTC é rima toante e consoante.)}
}
\end{tabular}
}
\label{poema1}
\end{table*}

\begin{table*}[ht]
\caption{As quatro sentenças candidatas ao verso 2 com os maiores escores.}
\centering
\resizebox{\columnwidth}{!}{%

\begin{tabular}{|l|l|l|c|c|c|c|c|c|}
\hline
Escansão                                                & Metro       & Tônicas           & ER         & ST             & AC         & RI         & RTC        & Escore \\ \hline
Des/lum/br\#a/vaas/ a/\#in/da/ o O/ri/\#en/te.     & 10          & 3, 6, 10       & 1      & 0              & 1          & 0          & 1          & 0.6          \\ \hline
\textbf{da/ Fa/v\#e/la/ ba/t\#i/am/nos/ de/ fr\#en/te.} & \textbf{10} & \textbf{3, 6, 10} & \textbf{1} & \textbf{0.333} & \textbf{1} & \textbf{0} & \textbf{1} & \textbf{0.667} \\ \hline
A/ ba/t\#a/lha/ pa/re/c\#ia/ i/mi/n\#en/te.           & 10          & 3, 7, 10          & 0.5        & 0          & 1          & 0          & 1          & 0.5          \\ \hline
E o/ d\#ia/ de/ri/v\#o/u/ tran/qui/la/m\#en/te.         & 10          & 2, 6, 9, 10       & 0.2        & 0.333          & 1          & 0          & 1          & 0.507          \\ \hline
\end{tabular}
}
\label{tab:poema1verso2}
\end{table*}

\begin{table*}[!ht]
\caption{Escore das quatro sentenças candidatas para ser o terceiro verso.}
\centering
\resizebox{\columnwidth}{!}{%
\begin{tabular}{|l|l|l|c|c|c|c|}
\hline
Escansão                                                     & Metro       & Tônicas              & ER           & ST         & RI             & Escore \\ \hline
Pa/r\#e/ce/ di/mi/nu/\#ir/ de al/ti/t\#u/de.                 & 10          & 2, 7, 10             & 0.2          & 0          & 0              & 0.067          \\ \hline
A/ tr\#o/pa/ com/ba/l\#i/da a/ba/l\#ou à/ t\#ar/de.          & 10          & 2, 6, 9, 10          & 0.4          & 0          & 0.91           & 0.164          \\ \hline
\textbf{Qu\#a/se/ t\#u/do/ es/t\#á/ s\#e/co/ de/ s\#e/de...} & \textbf{10} & \textbf{1, 3, 6, 10} & \textbf{0.6} & \textbf{0} & \textbf{0.273} & \textbf{0.291} \\ \hline
A/li/m\#en/ta/o e/ mi/t\#i/ga/lhe a/ s\#e/de.                & 10          & 3, 7, 10             & 0.5          & 0.25       & 0              & 0.25           \\ \hline
\end{tabular}
}
\label{tab:poema1verso3}
\vspace{-15pt}
\end{table*}

O primeiro verso do poema 1 (Tabela  \ref{poema1}) foi sorteado entre 24 sentenças metrificadas do grupo que possue a última sílaba ``te" e 10 sílabas poéticas. Por outro lado, o segundo verso foi selecionado com base no maior escore entre as sentenças do mesmo grupo, com o mesmo metro, mas que não finalizavam com a palavra 'febrilmente', a mesma palavra final do primeiro verso. A Tabela \ref{tab:poema1verso2} mostra as quatro sentenças com maior escore que foram candidatas ao segundo verso. A sentença escolhida como verso está em negrito.

A escolha do terceiro e quarto verso também usa a sentença com maior escore. O segundo grupo de sentenças sorteado foi com última sílaba igual a ``de", o qual possui somente quatro sentenças métricas disponíveis com metro 10.  As sentenças e seus escores estão na Tabela \ref{tab:poema1verso3}, com o segmento escolhido para ser o verso em negrito. Nessa tabela temos apenas três critérios, pois esse verso não possui um verso que rima já escolhido para ser a referência. 

Considerando a Tabela \ref{tab:poema1verso3}, para o verso quatro só restou as sentenças da primeira e da segunda linha. Isso porque a última palavra da sentença da linha quatro é igual a última palavra da sentença selecionada para ser o terceiro verso, portanto ela não poderia ser considerada. Tomando o verso três como referência, a sentença da linha dois acabou obtendo o maior escore, sendo então a escolhida como o último verso do poema.

O primeiro poema foi avaliado pelo sistema com um escore de 0.415, resultante da médias dos escores de cada verso (0.667, 0.291 e 0.287). Na Tabela \ref{poema1}, também é apresentado os critérios rítmicos global do poema, calculados a partir da média dos critérios dos versos.

No poema 2 (Tabela \ref{poema2}) é possível conferir os efeitos da alteração dos pesos dos  critérios na construção do poema. Enquanto no poema 1 todos os critérios tinham pesos iguais, no poema 2 o esquema rítmico e a rima consoante e toante passaram a ter seus pesos iguai a zero, enquanto o peso da rima interna dobrou. Como resultado o segundo verso do poema 2 mudou em relação ao primeiro. Essa mudança aconteceu, pois o escore do segundo verso no primeiro poema, usando os pesos do segundo poema é 0.333. Enquanto isso, escore no verso selecionado no segundo poema é de 0.341, valor maior do que o do verso escolhido no primeiro poema. O restante do poema seguiu da mesma forma, porém, como o escore de um dos versos mudou, consequentemente o escore geral do poema também foi alterado para 0.331.

\begin{table*}[ht]
\caption{Segundo Poema Gerado Pelo PROPOE 2.}
\centering
\resizebox{\columnwidth}{!}{%
\begin{tabular}{lllcccc}
\hline
\multicolumn{7}{l}{\multirow{4}{*}{\begin{tabular}[c]{@{}l@{}}Renovou-se a investida febrilmente\\ \textbf{Desce por ali a guarda da frente.}\\ Quase tudo está seco de sede\\ A tropa combalida abalou à tarde\end{tabular}}}                                                                                            \\
\multicolumn{7}{l}{}                                                                                                                                                                                                                                                                                             \\
\multicolumn{7}{l}{}                                                                                                                                                                                                                                                                                             \\
\multicolumn{7}{l}{}                                                                                                                                                                                                                                                                                             \\ \hline
\multicolumn{1}{|l|}{Escansão}                                                     & \multicolumn{1}{l|}{Metro}       & \multicolumn{1}{l|}{Tônicas}             & \multicolumn{1}{c|}{ST}         & \multicolumn{1}{c|}{AC}         & \multicolumn{1}{c|}{RI}             & \multicolumn{1}{c|}{Escore} \\ \hline
\multicolumn{1}{|l|}{Re/no/v\#ou/se a in/ves/t\#i/da/ fe/bril/m\#en/te.}           & \multicolumn{1}{l|}{10}          & \multicolumn{1}{l|}{3, 6, 10}            & \multicolumn{1}{c|}{-}          & \multicolumn{1}{c|}{-}          & \multicolumn{1}{c|}{-}              & \multicolumn{1}{c|}{-}              \\ \hline
\multicolumn{1}{|l|}{\textbf{D\#es/ce/ por/ a/l\#i/ a/ gu\#ar/da/ da/ fr\#en/te.}} & \multicolumn{1}{l|}{\textbf{10}} & \multicolumn{1}{l|}{\textbf{1, 5, 7 10}} & \multicolumn{1}{c|}{\textbf{0}} & \multicolumn{1}{c|}{\textbf{1}} & \multicolumn{1}{c|}{\textbf{0.182}} & \multicolumn{1}{c|}{\textbf{0.341}} \\ \hline
\multicolumn{1}{|l|}{Qu\#a/se/ t\#u/do/ es/t\#á/ s\#e/co/ de/ s\#e/de...}          & \multicolumn{1}{l|}{10}          & \multicolumn{1}{l|}{1, 3, 6, 7, 10}      & \multicolumn{1}{c|}{0}          & \multicolumn{1}{c|}{-}          & \multicolumn{1}{c|}{0.273}          & \multicolumn{1}{c|}{0.182}          \\ \hline
\multicolumn{1}{|l|}{A/ tr\#o/pa/ com/ba/l\#i/da a/ba/l\#ou à/ t\#ar/de.}          & \multicolumn{1}{l|}{10}          & \multicolumn{1}{l|}{2, 6, 9, 10}         & \multicolumn{1}{c|}{0}          & \multicolumn{1}{c|}{1}          & \multicolumn{1}{c|}{0.091}          & \multicolumn{1}{c|}{0.295}          \\ \hline
\multicolumn{3}{|c|}{Avaliação}                                                                                                                                  & \multicolumn{1}{c|}{0}          & \multicolumn{1}{c|}{1}          & \multicolumn{1}{c|}{0.182}          & \multicolumn{1}{c|}{0.331}          \\ \hline
\multicolumn{7}{l}{
\footnotesize{ (Os critérios com pesos zerados não foram adicionados a tabela (ver Tabela \ref{parametros}).)}
}
\end{tabular}
}
\label{poema2}
\end{table*}

\begin{table*}[ht]
\caption{Terceiro poema gerado pelo PROPOE 2.}
\centering
\resizebox{\columnwidth}{!}{%
\begin{tabular}{lllcccccc}
\hline
\multicolumn{9}{l}{\multirow{4}{*}{\begin{tabular}[c]{@{}l@{}}Haver refluído sobre si mesmo\\ Era o último, naquele rumo\\ Não foi uma carga, foi um bote\\ A formação brasileira no norte\end{tabular}}}                                                                                                                       \\
\multicolumn{9}{l}{}                                                                                                                                                                                                                                                                                                            \\
\multicolumn{9}{l}{}                                                                                                                                                                                                                                                                                                            \\
\multicolumn{9}{l}{}                                                                                                                                                                                                                                                                                                            \\ \hline
\multicolumn{1}{|l|}{Escansão}                                             & \multicolumn{1}{l|}{Metro} & \multicolumn{1}{l|}{Tônicas}          & \multicolumn{1}{c|}{ER}    & \multicolumn{1}{c|}{ST} & \multicolumn{1}{c|}{AC} & \multicolumn{1}{c|}{RI}    & \multicolumn{1}{c|}{RTC}  & \multicolumn{1}{c|}{Escore} \\ \hline
\multicolumn{1}{|l|}{ha/v\#er/ re/flu/\#í/do/ s\#o/bre/ s\#i/ m\#es/mo.}   & \multicolumn{1}{l|}{10}    & \multicolumn{1}{l|}{2, 5, 7, 9, 10}   & \multicolumn{1}{c|}{-}     & \multicolumn{1}{c|}{-}  & \multicolumn{1}{c|}{-}  & \multicolumn{1}{c|}{-}     & \multicolumn{1}{c|}{-}    & \multicolumn{1}{c|}{-}              \\ \hline
\multicolumn{1}{|l|}{\#E/ra/ o \#úl/ti/mo,/ na/qu\#e/le/ r\#u/mo.}         & \multicolumn{1}{l|}{9}     & \multicolumn{1}{l|}{1, 3, 7, 9}       & \multicolumn{1}{c|}{0.286} & \multicolumn{1}{c|}{0}  & \multicolumn{1}{c|}{1}  & \multicolumn{1}{c|}{0.1}   & \multicolumn{1}{c|}{0}    & \multicolumn{1}{c|}{0.227}          \\ \hline
\multicolumn{1}{|l|}{N\#ão/ f\#oi/ \#u/ma/ c\#ar/ga,/ f\#oi/ um/ b\#o/te.} & \multicolumn{1}{l|}{9}     & \multicolumn{1}{l|}{1, 2, 3, 5, 7, 9} & \multicolumn{1}{c|}{0.619} & \multicolumn{1}{c|}{0}  & \multicolumn{1}{c|}{-}  & \multicolumn{1}{c|}{0.1}   & \multicolumn{1}{c|}{-}    & \multicolumn{1}{c|}{0.24}           \\ \hline
\multicolumn{1}{|l|}{A/ for/ma/ç\#ão/ bra/si/l\#ei/ra/ no/ N\#or/te.}      & \multicolumn{1}{l|}{10}    & \multicolumn{1}{l|}{4, 7, 10}         & \multicolumn{1}{c|}{0.229} & \multicolumn{1}{c|}{0}  & \multicolumn{1}{c|}{1}  & \multicolumn{1}{c|}{0}     & \multicolumn{1}{c|}{0.5}  & \multicolumn{1}{c|}{0.346}          \\ \hline
\multicolumn{3}{|c|}{Avaliação}                                                                                                                 & \multicolumn{1}{c|}{0.378} & \multicolumn{1}{c|}{0}  & \multicolumn{1}{c|}{1}  & \multicolumn{1}{c|}{0.067} & \multicolumn{1}{c|}{0.25} & \multicolumn{1}{c|}{0.288}          \\ \hline
\end{tabular}
}
\label{poema3}
\end{table*}

\begin{table*}[!ht]
\caption{Quarto poema gerado pelo PROPOE 2.}
\centering
\resizebox{\columnwidth}{!}{%
\begin{tabular}{lllcccccc}
\hline
\multicolumn{9}{l}{\multirow{4}{*}{\begin{tabular}[c]{@{}l@{}}A vitória viria por si mesma\\ O aspecto reduzia-lhe a fama\\ Volviam à impaciência heróica\\ Di-lo uma comparação histórica\end{tabular}}}                                                                                                               \\
\multicolumn{9}{l}{}                                                                                                                                                                                                                                                                                                    \\
\multicolumn{9}{l}{}                                                                                                                                                                                                                                                                                                    \\
\multicolumn{9}{l}{}                                                                                                                                                                                                                                                                                                    \\ \hline
\multicolumn{1}{|l|}{Escansão}                                          & \multicolumn{1}{l|}{Metro} & \multicolumn{1}{l|}{Tônicas}     & \multicolumn{1}{c|}{ER}    & \multicolumn{1}{c|}{ST}    & \multicolumn{1}{c|}{AC} & \multicolumn{1}{c|}{RI} & \multicolumn{1}{c|}{RTC}  & \multicolumn{1}{c|}{Escore} \\ \hline
\multicolumn{1}{|l|}{A/ vi/t\#ó/ri/a/ vi/r\#ia/ por/ s\#i/ m\#es/ma.}   & \multicolumn{1}{l|}{10}    & \multicolumn{1}{l|}{3, 7, 9, 10} & \multicolumn{1}{c|}{-}     & \multicolumn{1}{c|}{-}     & \multicolumn{1}{c|}{-}  & \multicolumn{1}{c|}{-}  & \multicolumn{1}{c|}{-}    & \multicolumn{1}{c|}{-}              \\ \hline
\multicolumn{1}{|l|}{O/ as/p\#ec/to/ re/du/z\#ia/lhe a/ f\#a/ma.}       & \multicolumn{1}{l|}{10}    & \multicolumn{1}{l|}{3, 7, 10}    & \multicolumn{1}{c|}{0.75}  & \multicolumn{1}{c|}{0}     & \multicolumn{1}{c|}{1}  & \multicolumn{1}{c|}{0}  & \multicolumn{1}{c|}{0}    & \multicolumn{1}{c|}{0.35}           \\ \hline
\multicolumn{1}{|l|}{Vol/v\#i/am/ à im/pa/ci/\#ên/cia he/r\#ó/i/ca.}    & \multicolumn{1}{l|}{10}    & \multicolumn{1}{l|}{2, 7, 10}    & \multicolumn{1}{c|}{0.45}  & \multicolumn{1}{c|}{0}     & \multicolumn{1}{c|}{-}  & \multicolumn{1}{c|}{0}  & \multicolumn{1}{c|}{-}    & \multicolumn{1}{c|}{0.15}           \\ \hline
\multicolumn{1}{|l|}{D\#i/lo/ \#u/ma/ com/pa/ra/ç\#ã/o his/t\#ó/ri/ca.} & \multicolumn{1}{l|}{10}    & \multicolumn{1}{l|}{1, 3, 8, 10} & \multicolumn{1}{c|}{0.167} & \multicolumn{1}{c|}{0.062} & \multicolumn{1}{c|}{1}  & \multicolumn{1}{c|}{0}  & \multicolumn{1}{c|}{0.5}  & \multicolumn{1}{c|}{0.298}          \\ \hline
\multicolumn{3}{|c|}{Avaliação}                                                                                                         & \multicolumn{1}{c|}{0.456} & \multicolumn{1}{c|}{0.021} & \multicolumn{1}{c|}{1}  & \multicolumn{1}{c|}{0}  & \multicolumn{1}{c|}{0.25} & \multicolumn{1}{c|}{0.282}          \\ \hline
\end{tabular}
}
\label{poema4}
\vspace{-15pt}
\end{table*}

\begin{table*}[htb]
\caption{Soneto gerado pelo PROPOE 2.}
\centering
\resizebox{\columnwidth}{!}{%
\begin{tabular}{lllcccccc}
\hline
\multicolumn{9}{l}{\multirow{17}{*}{\begin{tabular}[c]{@{}l@{}}Sucedem-se meses e anos ardentes\\ É uma diátese e é uma síntese\\ Vivia-se à aventura, de expedientes\\ Hipóteses sobre a sua gênese\\ \\ Desceram ruidosamente as vertentes\\ Os adversários acotovelavam-se\\ Este é um rio sem afluentes\\ O contemplativo, então, levanta-se\\ \\ Foi, sem maior exame, aprovado\\ Ora, este fato era um aviso\\ Nada referia sobre o passado\\ \\ Tinha meio caminho andado\\ Guarda-a como capital precioso\\ Era, certo, o inimigo anelado\end{tabular}}} \\
\multicolumn{9}{l}{}                                                                                                                                                                                                                                                                                                                                                                                                                                                                                                                                             \\
\multicolumn{9}{l}{}                                                                                                                                                                                                                                                                                                                                                                                                                                                                                                                                             \\
\multicolumn{9}{l}{}                                                                                                                                                                                                                                                                                                                                                                                                                                                                                                                                             \\
\multicolumn{9}{l}{}                                                                                                                                                                                                                                                                                                                                                                                                                                                                                                                                             \\
\multicolumn{9}{l}{}                                                                                                                                                                                                                                                                                                                                                                                                                                                                                                                                             \\
\multicolumn{9}{l}{}                                                                                                                                                                                                                                                                                                                                                                                                                                                                                                                                             \\
\multicolumn{9}{l}{}                                                                                                                                                                                                                                                                                                                                                                                                                                                                                                                                             \\
\multicolumn{9}{l}{}                                                                                                                                                                                                                                                                                                                                                                                                                                                                                                                                             \\
\multicolumn{9}{l}{}                                                                                                                                                                                                                                                                                                                                                                                                                                                                                                                                             \\
\multicolumn{9}{l}{}                                                                                                                                                                                                                                                                                                                                                                                                                                                                                                                                             \\
\multicolumn{9}{l}{}                                                                                                                                                                                                                                                                                                                                                                                                                                                                                                                                             \\
\multicolumn{9}{l}{}                                                                                                                                                                                                                                                                                                                                                                                                                                                                                                                                             \\
\multicolumn{9}{l}{}                                                                                                                                                                                                                                                                                                                                                                                                                                                                                                                                             \\
\multicolumn{9}{l}{}                                                                                                                                                                                                                                                                                                                                                                                                                                                                                                                                             \\
\multicolumn{9}{l}{}                                                                                                                                                                                                                                                                                                                                                                                                                                                                                                                                             \\
\multicolumn{9}{l}{}                                                                                                                                                                                                                                                                                                                                                                                                                                                                                                                                             \\ \hline
\multicolumn{1}{|l|}{Escansão}                                                                     & \multicolumn{1}{l|}{Metro}                          & \multicolumn{1}{l|}{Tônicas}                                 & \multicolumn{1}{c|}{ER}                             & \multicolumn{1}{c|}{ST}                             & \multicolumn{1}{c|}{AC}                          & \multicolumn{1}{c|}{RI}                             & \multicolumn{1}{c|}{RTC}                            & \multicolumn{1}{c|}{Escore}                         \\ \hline
\multicolumn{1}{|l|}{Su/c\#e/dem/se/ m\#e/ses/ e \#a/nos/ ar/d\#en/tes.}                           & \multicolumn{1}{l|}{10}                             & \multicolumn{1}{l|}{2, 5, 7, 10}                             & \multicolumn{1}{c|}{-}                              & \multicolumn{1}{c|}{-}                              & \multicolumn{1}{c|}{-}                           & \multicolumn{1}{c|}{-}                              & \multicolumn{1}{c|}{-}                              & \multicolumn{1}{c|}{-}                                      \\ \hline
\multicolumn{1}{|l|}{é/ \#u/ma/ di/\#á/te/se/ e é \#u/ma/ s\#ín/te/se.}                            & \multicolumn{1}{l|}{10}                             & \multicolumn{1}{l|}{2, 5, 8, 10}                             & \multicolumn{1}{c|}{0.6}                            & \multicolumn{1}{c|}{0}                              & \multicolumn{1}{c|}{-}                           & \multicolumn{1}{c|}{0.25}                           & \multicolumn{1}{c|}{-}                              & \multicolumn{1}{c|}{0.283}                                  \\ \hline
\multicolumn{1}{|l|}{Vi/v\#ia/se à a/ven/t\#u/ra,/ de ex/pe/di/\#en/tes.}                          & \multicolumn{1}{l|}{10}                             & \multicolumn{1}{l|}{2, 5, 10}                                & \multicolumn{1}{c|}{0.75}                           & \multicolumn{1}{c|}{0}                              & \multicolumn{1}{c|}{1}                           & \multicolumn{1}{c|}{0}                              & \multicolumn{1}{c|}{1}                              & \multicolumn{1}{c|}{0.55}                                   \\ \hline
\multicolumn{1}{|l|}{Hi/p\#ó/te/ses/ s\#o/bre/ a/ s\#u/a/ g\#ê/ne/se.}                             & \multicolumn{1}{l|}{10}                             & \multicolumn{1}{l|}{2, 5, 8, 10}                             & \multicolumn{1}{c|}{0.675}                          & \multicolumn{1}{c|}{0}                              & \multicolumn{1}{c|}{1}                           & \multicolumn{1}{c|}{0.083}                          & \multicolumn{1}{c|}{0}                              & \multicolumn{1}{c|}{0.352}                                  \\ \hline
\multicolumn{1}{|l|}{Des/c\#e/ram/ r\#ui/do/sa/men/te As/ ver/t\#en/tes.}                          & \multicolumn{1}{l|}{10}                             & \multicolumn{1}{l|}{2, 4, 10}                                & \multicolumn{1}{c|}{-}                              & \multicolumn{1}{c|}{-}                              & \multicolumn{1}{c|}{-}                           & \multicolumn{1}{c|}{-}                              & \multicolumn{1}{c|}{-}                              & \multicolumn{1}{c|}{-}                                      \\ \hline
\multicolumn{1}{|l|}{Os/ ad/ver/s\#á/rios/ a/co/to/ve/l\#a/vam/se.}                                & \multicolumn{1}{l|}{10}                             & \multicolumn{1}{l|}{4, 10}                                   & \multicolumn{1}{c|}{0.667}                          & \multicolumn{1}{c|}{0}                              & \multicolumn{1}{c|}{1}                           & \multicolumn{1}{c|}{0}                              & \multicolumn{1}{c|}{0}                              & \multicolumn{1}{c|}{0.333}                                  \\ \hline
\multicolumn{1}{|l|}{\#Es/te/ é/ um/ r\#i/o/ Sem/ a/flu/\#en/tes.}                                 & \multicolumn{1}{l|}{10}                             & \multicolumn{1}{l|}{1, 5, 10}                                & \multicolumn{1}{c|}{0.225}                          & \multicolumn{1}{c|}{0}                              & \multicolumn{1}{c|}{1}                           & \multicolumn{1}{c|}{0}                              & \multicolumn{1}{c|}{1}                              & \multicolumn{1}{c|}{0.445}                                  \\ \hline
\multicolumn{1}{|l|}{O/ con/tem/pla/t\#i/vo,/ en/t\#ão,/ le/v\#an/ta/se.}                          & \multicolumn{1}{l|}{10}                             & \multicolumn{1}{l|}{5, 8, 10}                                & \multicolumn{1}{c|}{0.35}                           & \multicolumn{1}{c|}{0}                              & \multicolumn{1}{c|}{1}                           & \multicolumn{1}{c|}{0}                              & \multicolumn{1}{c|}{0.5}                            & \multicolumn{1}{c|}{0.37}                                   \\ \hline
\multicolumn{1}{|l|}{F\#oi,/ Sem/ ma/i/\#or/ e/x\#a/me, a/pro/v\#a/do.}                            & \multicolumn{1}{l|}{10}                             & \multicolumn{1}{l|}{1, 5, 7, 10}                             & \multicolumn{1}{c|}{-}                              & \multicolumn{1}{c|}{-}                              & \multicolumn{1}{c|}{-}                           & \multicolumn{1}{c|}{-}                              & \multicolumn{1}{c|}{-}                              & \multicolumn{1}{c|}{-}                                      \\ \hline
\multicolumn{1}{|l|}{\#O/ra,/ \#es/te/ f\#a/to/ \#e/ra um/ a/v\#i/so.}                             & \multicolumn{1}{l|}{10}                             & \multicolumn{1}{l|}{1, 3, 5, 7, 10}                          & \multicolumn{1}{c|}{0.8}                            & \multicolumn{1}{c|}{0}                              & \multicolumn{1}{c|}{-}                           & \multicolumn{1}{c|}{0}                              & \multicolumn{1}{c|}{-}                              & \multicolumn{1}{c|}{0.267}                                  \\ \hline
\multicolumn{1}{|l|}{N\#a/da/ re/fe/r\#i/a/ s\#o/bre o/ pa/ss\#a/do.}                              & \multicolumn{1}{l|}{10}                             & \multicolumn{1}{l|}{1, 5, 7, 10}                             & \multicolumn{1}{c|}{0.9}                            & \multicolumn{1}{c|}{0}                              & \multicolumn{1}{c|}{1}                           & \multicolumn{1}{c|}{0}                              & \multicolumn{1}{c|}{1}                              & \multicolumn{1}{c|}{0.58}                                   \\ \hline
\multicolumn{1}{|l|}{T\#i/nha/ m\#e/i/o/ ca/m\#i/nho/ an/d\#a/do.}                                 & \multicolumn{1}{l|}{10}                             & \multicolumn{1}{l|}{1, 3, 7, 10}                             & \multicolumn{1}{c|}{-}                              & \multicolumn{1}{c|}{-}                              & \multicolumn{1}{c|}{-}                           & \multicolumn{1}{c|}{-}                              & \multicolumn{1}{c|}{-}                              & \multicolumn{1}{c|}{-}                                      \\ \hline
\multicolumn{1}{|l|}{Gu\#ar/daa/ c\#o/mo/ ca/pi/t\#al/ pre/ci/\#o/so.}                             & \multicolumn{1}{l|}{10}                             & \multicolumn{1}{l|}{1, 3, 7, 10}                             & \multicolumn{1}{c|}{1}                              & \multicolumn{1}{c|}{0}                              & \multicolumn{1}{c|}{1}                           & \multicolumn{1}{c|}{0}                              & \multicolumn{1}{c|}{0}                              & \multicolumn{1}{c|}{0.4}                                    \\ \hline
\multicolumn{1}{|l|}{\#E/ra,/ c\#er/to,/ o i/ni/m\#i/go a/ne/l\#a/do.}                             & \multicolumn{1}{l|}{10}                             & \multicolumn{1}{l|}{1, 3, 7, 10}                             & \multicolumn{1}{c|}{1}                              & \multicolumn{1}{c|}{0.125}                          & \multicolumn{1}{c|}{1}                           & \multicolumn{1}{c|}{0}                              & \multicolumn{1}{c|}{1}                              & \multicolumn{1}{c|}{0.625}                                  \\ \hline
\multicolumn{3}{|c|}{Avaliação}                                                                                                                                                                                         & \multicolumn{1}{c|}{0.697}                          & \multicolumn{1}{c|}{0.013}                          & \multicolumn{1}{c|}{1}                           & \multicolumn{1}{c|}{0.033}                          & \multicolumn{1}{c|}{0.562}                          & \multicolumn{1}{c|}{0.42}                                   \\ \hline
\end{tabular}
}
\label{poema5}
\end{table*}

As entradas do poema 3 se diferem do poema 1 apenas pelo metro (Tabela  \ref{parametros}). Enquanto no poema 1 o metro de todos os versos é 10, no poema 3 o metro do primeiro e quarto verso é 10 e do segundo e terceiro é 9. Embora a estratégia da semente aleatória tenha se mantido igual às dos poemas 1 e 2, a rima mudou no poema 3 (Tabela  \ref{poema3}). Essa mudança foi consequência da alteração do metro, pois as opções de grupos de sentenças possíveis para serem sorteados foi modificado em comparação com os poemas anteriores.

O processo de construção do poema 3 é semelhante ao do poema 1 e 2, exceto pelo fato de que os candidados aos versos dois e três foram sentenças com metro 9. O poema final foi avaliado com escore 0.288.

No poema 4 (Tabela  \ref{poema4}) gerou-se um poema com as mesmas entradas do poema 1, mas removendo a semente aleatória. Como esperado a rima mudou completamente, pois agora outros grupos de sentença foram sorteados. O escore do novo poema foi avaliado em 0.282.
 
O soneto, apresentado na Tabela \ref{poema5} foi quinto poema gerado. Soneto é uma estrutura clássica de poema estruturado com quatro estrofes, sendo as duas primeiros formado por quatro versos e os dois últimos formados por três versos. Esse tipo de poema geralmente apresenta versos de dez ou doze
sílabas, e as rimas dos dois primeiros versos são diferentes das rimas dos dois últimos \cite{goldstein1985versos}. Considerando esses requisitos montou-se os parâmetros de entrada do poema 5 apresentados na Tabela \ref{parametros}.

Como cada estrofe começa com uma sentença aleatória, esses versos não tiverem seus critérios calculados. Os critérios de acentuação da rima e rima toante e consoante também não são calculados para os versos que apresentam um padrão de rima pela primeira vez. Dessa forma, o segundo e o décimo verso do soneto não tiveram os dois critérios calculados, mas o sexto verso, que é o primeiro verso com essa rima na segundo estrofe, teve os critérios calculados, usando o último verso da estrofe anterior como referência. No final, o poema é avaliado pela média do escore de cada verso. O que resultou em um escore de 0.42 para o soneto.

\section{Conclusão}

O PROPOE é uma ferramenta de geração automática de poemas com base em sentenças metrificadas extraídas de obras literárias brasileiras. Nessa segunda versão, buscou-se ampliar as possibilidades de estrutura de poemas e combinação de efeitos sonoros em comparação com a versão original. Passou-se a permitir a definição de metros diferentes para cada verso, foi adicionada pontuação para sílabas tônicas que ocorrem em diferentes posições dos versos, inseriram-se os pesos dos critérios como parâmetro de entrada, além da inclusão dos critérios de rima interna, e rima toante e consoante. A incorporação dos pesos dos critérios aos parâmetros de entrada dá ao usuário mais controle sobre o poema que ele deseja gerar, admitindo a exploração de novos ritmos e efeitos sonoros com certa precisão.

Nas versões seguintes, ainda seria possível incluir mais critérios rítmicos como aliteração e assonância. A abordagem para montagem do poema poderia evoluir para alguma que, em vez de realizar somente otimizações locais dos critérios rítmicos, também se preocupasse em realizar uma otimização global. No PROPOE 2, assim como na versão original, não foi implementada nenhuma restrição semântica, deixando o contexto e a coesão temática a cargo da seleção de sentenças de uma única obra. A agregação de uma dimensão semântica à ferramenta poderia gerar poemas mais coesos, além de permitir a seleção de sentenças em um contexto maior usando múltiplas obras literárias.

\bibliographystyle{sbc}
\bibliography{sbc-template}

\end{document}